\documentclass[10pt,twocolumn,letterpaper]{article}

\usepackage{iccv}
\usepackage{times}
\usepackage{epsfig}
\usepackage{graphicx}
\usepackage{amsmath}
\usepackage{amssymb}
\usepackage{makecell}
\usepackage{float}
\usepackage{caption}
\usepackage{subcaption}
\usepackage{lipsum}  
\usepackage{algorithm}  
\usepackage{algpseudocode}
\usepackage{pifont}
\usepackage{comment}
\usepackage{xcolor}
\usepackage{booktabs}

\usepackage{color, colortbl}

\usepackage{multirow}
\usepackage[normalem]{ulem}
\useunder{\uline}{\ul}{}

\definecolor{LightBlue}{rgb}{0.95,0.96,1}

\usepackage[pagebackref=true,breaklinks=true,letterpaper=true,colorlinks,bookmarks=false]{hyperref}

\iccvfinalcopy 

\def\iccvPaperID{****} 

\ificcvfinal\fi

\begin{document}

\title{Are current long-term video understanding datasets long-term?}



\author{Ombretta Strafforello\\
TU Delft, TNO\\ 
{\tt\small o.strafforello@tudelft.nl}
\and
Klamer Schutte\\
TNO\\
{\tt\small klamer.schutte@tno.nl}
\and
Jan van Gemert\\
TU Delft\\ 
{\tt\small j.c.vangemert@tudelft.nl}
}

\maketitle
\ificcvfinal\thispagestyle{empty}\fi

\begin{abstract}
Many real-world applications, from sport analysis to surveillance,  benefit from automatic long-term action recognition. 
In the current deep learning paradigm for automatic action recognition, it is imperative that models are trained and tested on datasets and tasks that evaluate if such models actually learn and reason over long-term information. 
In this work, we propose a method to evaluate how suitable a video dataset is to evaluate models for long-term action recognition.
To this end, we define a \textit{long-term action} as  excluding all the videos that can be correctly recognized using solely short-term information. We test this definition on existing long-term classification tasks on three popular real-world datasets, namely Breakfast, CrossTask and LVU, to determine if these datasets are truly evaluating long-term recognition. 
Our study reveals that these datasets can be effectively solved using shortcuts based on short-term information. 
Following this finding, we encourage long-term action recognition researchers to make use of datasets that need long-term information to be solved. 
\end{abstract}

\section{Introduction}
\label{sec:introduction}

\begin{figure*}[!ht]
\centering
\includegraphics[width=1\textwidth]{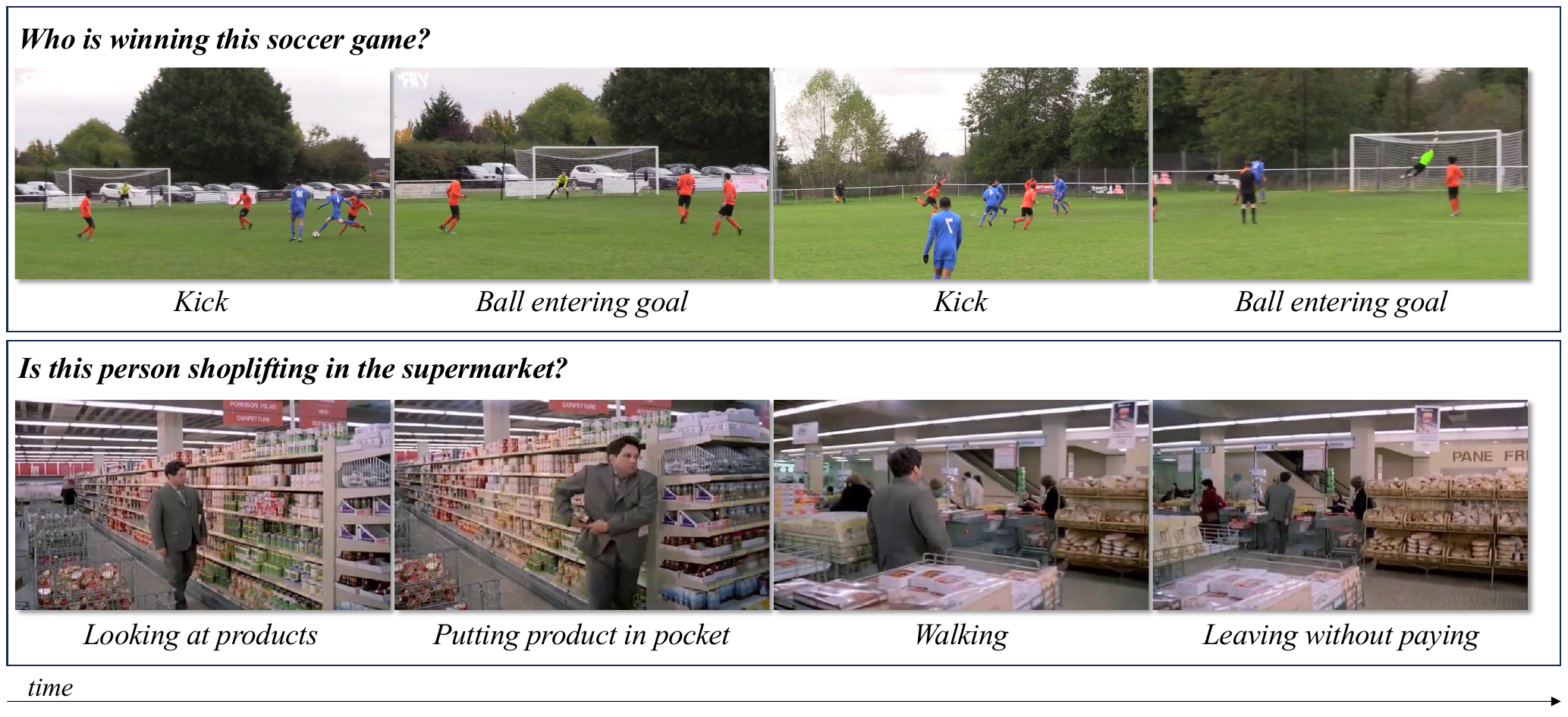}
\caption{Example of truly long-term actions. \textit{Top:} Who is winning this soccer game?
\textsuperscript{1}, \textit{Bottom:} Is this person shoplifting in the supermarket?\textsuperscript{2}. In both cases, it is not possible to answer correctly without considering multiple short-term actions together, their order and relations over time. To understand who is winning the soccer game, it is necessary to recognize and count the goals scored since the beginning of the game. 
To recognize shoplifting, it is not enough to see a person putting a product in their pocket: also the short-term action \textit{leaving without paying} needs to occur. \\
  \small\textsuperscript{1}Source: \href{https://www.youtube.com/watch?v=8MRHcLMQgvU}{YouTube};
  \small\textsuperscript{2}Source: \href{https://www.youtube.com/watch?v=5tVz7IChlWI}{YouTube} from movie \textit{Un povero ricco}, by Pasquale Festa Campanile (1983).}
\label{fig:fig1}
\end{figure*}

Many interesting actions happening in the real world are long-term. That is, they are composed of several short sub-actions, that we refer to as \textit{short-term actions}.
For an action to be \emph{long-term}, we deem that recognizing a single-short term action is not enough, and reasoning about the order and the relationship of short-term actions is required. 
Two examples of long-term actions, shown in Figure \ref{fig:fig1}, are \textit{winning a soccer game} and \textit{shoplifting in the supermarket}. To understand which team is winning a soccer game, it is necessary to recognize and count the goals scored since the beginning of the game. 
For the other example, recognizing if a person is shoplifting, it is necessary to observe a person storing a product in their pocket \textit{and} leaving the supermarket without paying. In both examples, it is not possible to recognize the actions without reasoning on multiple ordered short-term actions. 

\begin{figure*}[ht!]
\centering
\includegraphics[width=1.0\textwidth]{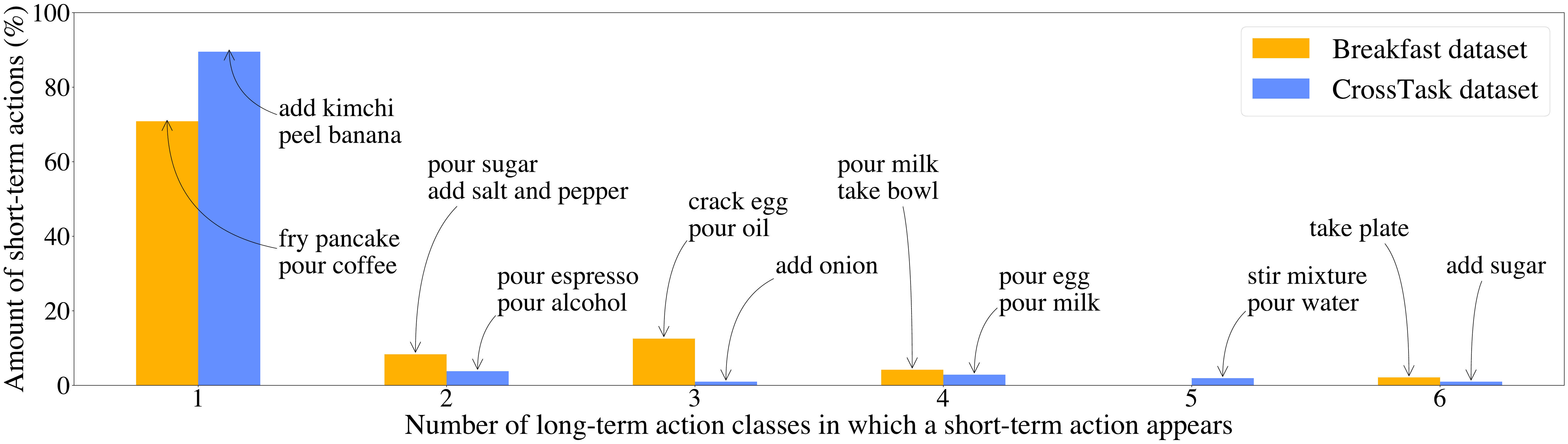}
\caption{We analyze two popular long-term datasets with long-term and short-term action annotations, Breakfast (coarse annotations) \cite{kuehne2014language} and CrossTask \cite{zhukov2019cross} (primary tasks). We count in how many long-term actions the short-term action appears. 
Recurrent short-term actions, like \textit{pour milk} and \textit{pour egg} appear in four different long-term action classes. More specific short-term actions, like \textit{fry pancake} and \textit{add kimchi}, only occur in one long-term action class. 
We find that a large percentage of short-term actions (70.8\% for Breakfast and 89.5\% for CrossTask) appears only in one long-term action class. 
This implies that recognizing a single short-term action might be sufficient to correctly infer the long-term actions in these datasets.
}
\label{fig:unique_subactions}
\end{figure*}

Achieving automatic long-term action recognition is important because it can be used to solve real-world problems, from analyzing sports videos, to understanding movies and recognizing threats in surveillance footage. 
To make it possible, we need purpose-built computer vision models, that are trained and evaluated on datasets that need long-term reasoning to be solved.
While working on long-term action recognition, 
we notice that every video in the Breakfast dataset \cite{kuehne2014language}, a go-to choice in long-term video understanding research 
\cite{guo2022uncertainty, hussein2019timeception, li2022bridge, zhou2021graph}, contains short-term actions that map to a single long-term action. 
This implies that accurately recognizing a short-term action in a Breakfast video should be sufficient to infer the corresponding long-term action. 
We analyze the short-term actions of another popular instructional video dataset, CrossTask \cite{zhukov2019cross}, and find the same occurrence in 97.72\% of its primary tasks videos.
We illustrate our statistics on the short-term action occurrences in Figure \ref{fig:unique_subactions}.
Since deep learning models are known to use shortcuts to solve classification tasks \cite{geirhos2020shortcut}, the models trained and tested on these datasets might learn to exploit short-term information, without encoding any long-term relations. 

Motivated by this finding, we propose a method to diagnose whether a long-term dataset is suitable to study long-term action recognition, or can be solved using solely short-term information.
To this end, we define two requirements for an action to be long-term: 
(1) The action is \textit{recognizable only from multiple short-term actions} and not from a single short-term action. 
(2) The action maps to a \textit{single label}. 
The first requirement makes long-term action recognition impossible without reasoning over an extended time span. Models that lack this capability, for example based on straightforward pooling operations over time 
\cite{wang2016temporal}, cannot recognize long-term actions. 
The second requirement leads to discarding multi-label action recognition datasets, like Charades \cite{sigurdsson2016hollywood}, MultiTHUMOS \cite{yeung2018every} and EPIC-Kitchens \cite{damen2018scaling}, as long-term action datasets. In these datasets, the task is to recognize each short-term action contained in the videos. This task could be solved by classifying each short-term action one at a time, 
while here we are interested in the case where the classification can be made only after reasoning over multiple short-term actions together.

We design a user study to assess whether a video dataset contains long-term action videos 
that are not recognizable from a single short-term action. Our study is based on two surveys where users have to watch a video and predict the long-term action being performed in the video. In the \textit{Full Videos Survey}, the users can watch the full video, while in the \textit{Video Segments Survey} a separate group of users can watch only a single short clip extracted from the full video. 
We measure the average action recognition accuracy of the users per video for each survey. 
The \textit{Full Videos Survey} gives an upper bound to the user long-term action recognition performance. Comparing the accuracy obtained from the \textit{Video Segments Survey} to the upper bound gives an estimate of how many videos in the dataset require long-term information to be correctly recognized. If the action recognition performance of the two groups of users is close, we can conclude that most of the videos in the dataset are not suitable to train and evaluate models for long-term action recognition, because they can be recognized solely by exploiting short-term information.

We apply our proposed method to 
the aforementioned Breakfast and CrossTask datasets 
and to the Long-form Video Understanding benchmark (LVU) \cite{wu2021towards}, recently proposed for long-term video recognition tasks in movies. 
We implement the user studies on Amazon Mechanical Turk \cite{mturk} and collect responses from more than 150 users. 
Our results show that looking at a single short video segment is sufficient to recognize 90\% and 97.2\% of the analyzed videos from Breakfast and CrossTask. 
Similarly, we find that most of the content understanding tasks in LVU can be classified without long-term information, and that some video segments in this dataset are misclassified by users due to annotation noise.
We conclude that the aforementioned datasets might not be suitable to develop new methods for long-term action recognition in videos, because they can be solved by ignoring long-term information.
We recommend long-term video understanding researchers to be careful when using these datasets and encourage the community to collect more representative video datasets.

In summary, the contributions of our study can be outlined as follows:
(1) We provide a definition of long-term action datasets 
that 
should prevent long-term action recognition models to use traditional short-term action recognition as a shortcut to solve the task. 
(2) We introduce a method to investigate whether a video dataset 
meets this definition of long-term action.
(3) We find that short-term information is, in most cases, sufficient to solve long-term video understanding tasks in three commonly used datasets. Thus, we recommend against using these datasets in further research on long term action recognition models. 
The code and responses from our user study are publicly available\footnote{\url{https://github.com/ombretta/longterm_datasets}}. 

\section{Related work}
\subsection{Action recognition with deep learning}

The progress of deep learning (DL) has brought significant advancements 
in automatic action recognition.
DL-based models learn to extract discriminative spatial and temporal features directly from the RGB frames of the training videos.
Current action recognition models are composed of 3D convolutional networks \cite{ji20123d}, like I3D \cite{carreira2017quo}, C3D \cite{tran2015learning}, Slow-Fast \cite{feichtenhofer2019slowfast}. More recently, 
attention-based architectures have also shown competitive performance on action recognition tasks. Examples include ViViT \cite{arnab2021vivit}, TimeSformer \cite{bertasius2021space} and Video Swin Transformer \cite{liu2022video}.
%
%
When pre-trained on sufficiently large datasets, like Kinetics \cite{carreira2017quo} or ActivityNet \cite{caba2015activitynet}, these models can achieve state-of-the art action recognition on \textit{short} videos datasets, like UCF101 \cite{soomro2012ucf101}, HMDB51 \cite{kuehne2011hmdb} and Something-Something \cite{goyal2017something}. However, they are not suitable to learn long-term dynamics in long videos, either due to their limited temporal receptive field or the high computational requirements.

\subsection{Long-term action recognition}

Long-term action recognition refers to the task of recognizing and understanding human actions composed of several short-term actions, possibly involving multiple objects and movements \cite{zhou2021graph}. Examples include cooking a recipe \cite{kuehne2014language}, performing a medical surgery \cite{sharghi2020automatic} or playing a sport game \cite{yeung2018every}.
Usually, long-term actions require an extended period of time to be executed, e.g. above one minute \cite{hussein2019timeception}. 
Several works that tackled the problem of long-term action recognition use different names and definitions for the same concepts. In fact, long-term actions can also be referred to in the literature as \textit{long-range activities} \cite{
hussein2020timegate} or \textit{complex activities} \cite{guo2022uncertainty, hussein2019timeception}.
Being composed of multiple steps, the activities in \textit{instructional videos} share the same properties of long-term actions \cite{li2022bridge, miech2020end, zhou2018towards} and can be comprised into this category.
Finally, also \textit{long-form} video understanding involves reasoning over human-object interactions in long videos \cite{wu2021towards, yang2023relational} and can be considered as an 
instance of long-term action recognition.

Traditional DL-based action recognition models \cite{carreira2017quo, feichtenhofer2019slowfast, tran2015learning, wang2016temporal} are deemed insufficient to capture discriminative spatio-temporal features that encode long-term information and the semantic relations between the sub-actions. A variety of models have been proposed to overcome this limitation. Hussein \textit{et al.} \cite{hussein2019timeception} proposed to capture long-term information with multi-scale temporal convolution. Yu \textit{et al.} \cite{yu2020rhyrnn} used Recurrent Neural Networks to model long video sequences capturing temporal information at different rhythms.
Ballan \textit{et al.} \cite{ballan2021long} showed that explicitly focusing on the actor performing the long-term action improves the recognition performance. Different approaches 
showed that long-term action recognition can be tackled using graph-based representations, where the nodes correspond to short-term entities and the edges to their interaction over space and time \cite{hussein2019videograph, ji2020action, zhou2021graph}.
Finally, Transformer architectures have been designed to model long-term information in a compute- \cite{islam2022long, wu2022memvit} and data-efficient \cite{guo2022uncertainty} fashion.





Despite their success, DL-based action recognition models can find shortcuts in the data that let them solve action recognition without learning semantic features, for example classifying the action based on the background scene \cite{choi2019can, geirhos2020shortcut, xiao2021noise}. 
In this work, we try to address this problems by analyzing whether commonly used video datasets for long-term action recognition are representative for training DL models, or can be solved using short-term shortcuts. 

\subsection{Long-term video datasets}
Several datasets have been proposed in the literature to study long-term video understanding tasks. 
%
%
CATER \cite{girdhar2019cater} is an ideal example of a dataset that requires long-term information. It involves tracking geometrical shapes that move in a 3D space over time. Sometimes bigger shapes incorporate smaller shapes, rendering their localization impossible without continuous reasoning about past information. As a consequence, models that are not truly long-term fail on this dataset. Unfortunately, the CATER dataset is highly synthetic and cannot be used to train models for real-world applications. 

Real-world datasets mostly include cooking \cite{damen2018scaling, kuehne2014language, stein2013combining, zhou2018towards}, home activities \cite{sigurdsson2016hollywood, 7961782}, sports \cite{yeung2018every} and instructional videos \cite{alayrac2017learning, tang2019coin, zhou2018towards, zhukov2019cross}. A comprehensive overview of long-term video understanding datasets is provided in Table \ref{tab:long_term_datasets}.
Many of these datasets, for example Charades \cite{sigurdsson2016hollywood}, Epic Kitchens \cite{damen2018scaling} and MultiTHUMOS \cite{yeung2018every}, contain long videos annotated with fine-grained, short-term actions. They can be used for multi-label action recognition, where the task is to predict every short-term action occurring in the video, or for fine-grained action localization.
Differently, here we are interested in the single-label classification case, where a global label describes the long-term activity happening in the video. 
%
The single label should be recognizable only by reasoning over multiple short-term actions. 

Previous work showed that video datasets are sometimes biased towards appearance \cite{byvshev20223d} and better recognizable by short-term over long-term information \cite{videobagnet}.
Similarly, in this work we explore whether the global labels of datasets proposed for long-term video understanding tasks can be predicted without long-term information. 
We choose for our study three popular datasets that include single, video-level labels and cover different long-term dataset categories: Breakfast, CrossTask and LVU. Breakfast \cite{kuehne2014language} is a \textit{complex action recognition} dataset used in several works on long-term video understanding \cite{guo2022uncertainty, hussein2019timeception, li2022bridge, zhou2021graph}.  
CrossTask \cite{zhukov2019cross} is a dataset of \textit{instructional videos}, which are composed of several short-term steps that contribute to the completion of a long-term task.
Finally, the \textit{Long-form Video Understanding} (LVU) dataset \cite{wu2021towards} was proposed to learn complex long-term relationships, in contrast to short-term patterns, in video clips extracted from movies.

%
%



\begin{table}[]
\begin{tabular}{lllll}
\hline 
\textbf{Dataset} & \textbf{\#Videos} & \textbf{Length} & \textbf{\#L.T.} & \textbf{\#S.T.}  \\ \hline 

\rowcolor{LightBlue}
{\footnotesize COFFEE \cite{alayrac2017learning}} & 150 & 2 & 5 & 51 \\ 
\multirow{2}{*}{\shortstack[l]{{\footnotesize Epic-Kitchens \cite{damen2018scaling}}}}
 & \multirow{2}{*}{432} & \multirow{2}{*}{7.5} & \multirow{2}{*}{-} & 149,\\
 & & & & 323  \\ 
\rowcolor{LightBlue}
{\footnotesize Breakfast \cite{kuehne2014language}} & 2k & 2.3 & 10 & 48 \\ 
{\footnotesize Composite \cite{rohrbach2012script}} & 212 & 1-23 & 41 & 218 \\
\rowcolor{LightBlue}
{\footnotesize Charades \cite{sigurdsson2016hollywood}} & 10k & 0.5 & - & 157 \\ 
{\footnotesize 50-Salads \cite{stein2013combining}} & 54 & 6.4 & - & 17  \\ 
\rowcolor{LightBlue}
{\footnotesize COIN \cite{tang2019coin}} & 11.8k & 2.4 & 180 & 778 \\ 
{\footnotesize IKEA FA \cite{toyer2017human}} & 101 & 2-4 & - & 12 \\
\rowcolor{LightBlue}
{\footnotesize DAHLIA \cite{7961782}} & 51 & 39 & 7 & - \\ 
\multirow{3}{*}{\shortstack[l]{{\footnotesize LVU - Content} \\ {\footnotesize understanding \cite{wu2021towards}}}} & 226 & 1-3 & 4 & - \\ 
&  1.3k & 1-3 & 5 & - \\
&  723 & 1-3 & 6 & - \\
\rowcolor{LightBlue}
{\footnotesize Multi-THUMOS \cite{yeung2018every}} & 413 & 3 & - & 65 \\ 
{\footnotesize YouCookII \cite{zhou2018towards}} & 2k  & 5.3 & 89 & - \\ 
\rowcolor{LightBlue}
{\footnotesize CrossTask \cite{zhukov2019cross}} & 4.7k & 3-6 & 83 & 517 \\ 

\hline 
\end{tabular}
\caption{Overview of current real-world datasets proposed for long-term video understanding tasks. 
We report the (approximate) number of videos, the average video length in minutes, 
the number of global \textit{long-term} (L.T.) and \textit{short-term} (S.T.) 
action recognition classes, if it applies. 
}
\label{tab:long_term_datasets}
\vspace*{-2mm}
\end{table}

\section{Assessing long-term action recognition datasets}
\subsection{User study}
According to our definition, an action is long-term if it cannot be classified from a single 
short video segment. We design a user study to test whether current long-term video understanding datasets respect this property.
Our user study consists of two surveys. 
In the \textit{Full Videos Survey}, the users are presented with the full-length videos from the datasets. 
In the \textit{Video Segments Survey}, the users are presented with a short video segment extracted from a full-length video. 
%
In both surveys, the users are instructed to watch the video clip and express what action is being performed in the full video, 
in their opinion. 
The users are provided with a list of possible actions, which correspond to the classes from the analyzed long-term action datasets, and have to select exactly one action class from the list. We include the additional option ''\textit{I am not sure}'', to let the users express uncertainty when they are in doubt about which action to select.

From the collected user votes in the \textit{Full Videos Survey} and the \textit{Video Segments Survey}, we calculate and compare the action recognition accuracy. If the users from the two groups perform similarly, we can conclude that the videos do not contain long-term actions, as they can be recognized from single short-term actions comparably well than looking at the full videos.
We also calculate the user agreement per survey, measured with Krippendorff's $\alpha$ \cite{krippendorff2011computing}, which gives an indication of how subjective the prediction task is. We expect that the more a video is difficult to classify, the more subjective the choice will be, thus resulting in low agreement.

\subsection{Measuring 
recognition accuracy}


From the \textit{Full Videos Survey}, we collect user votes per class for each full-length video. In each full video, we express the votes in percentages ($\%user\_votes_v(c)$), which we obtain by dividing the votes per class by the amount of votes collected for the full video.
As formalized in Equation \ref{eq:pred_full_video}, given $\mathcal{C}$ classes from the evaluated dataset, excluding the \textit{I am not sure} option, we assign to the full video prediction ($pred(v)$) the class voted by the majority of the users. 
The long-term action recognition accuracy is given by the number of full videos assigned with the correct class over the number of full videos considered in the study for the dataset. 

\begin{equation}
   pred(v) = \underset{c\in \mathcal{C} 
   }{\arg\max } \quad \%user\_votes_v(c)
\label{eq:pred_full_video}
\end{equation}

\begin{figure*}[ht!]
\centering
\includegraphics[width=1\textwidth]{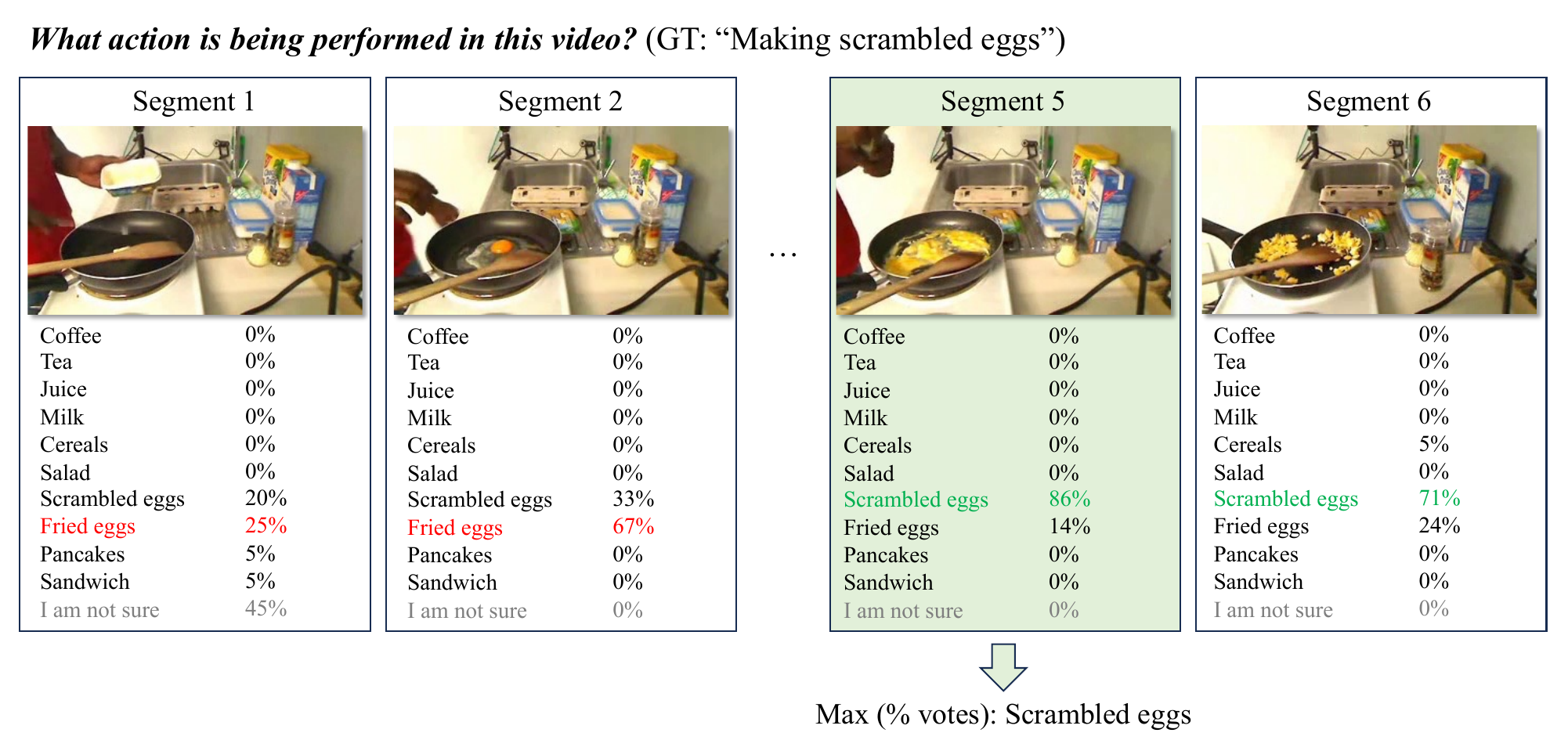}
\caption{In the \textit{Video Segments Survey}, users have to understand what is happening in a long video by looking only at one short segment. We ask the users to vote for a video class and obtain predictions per segment. We assign to the full video the segment prediction with the highest percentage of 
votes for one class. In the example, taken from the Breakfast dataset \cite{kuehne2014language}, \textit{Segment 5} determines the video prediction \textit{Scrambled eggs}.
}
\label{fig:method_segments}
\end{figure*}

In the \textit{Video Segments Survey}, we collect user votes for every segment $s_v$ in a full video. Again, for each segment we calculate the percentage of votes per class $\%user\_votes(c)$.
Then, we extract the full video prediction from the votes of a single segment. To do this, we select the segment $s^*_v$ with highest 
percentage of votes for 
a single class, excluding 
the \textit{I am not sure} option. 
This approach is formalized in Equation \ref{eq:pred_video_segments}.
In the example in Figure \ref{fig:method_segments}, the full video is assigned the class \textit{Making scrambled eggs}, which is voted by 
86\% of users in \textit{Segment 5}, which is the maximum ratio of votes for one class across the video segments. 
According to our definition, if the full-length video is long-term, there should be no video segments that lead to the right predicted class. 
The accuracy is given by the number of full videos assigned with the correct label over the number of full videos considered in the study. 

\begin{equation}\label{eq:pred_video_segments}
    pred(v) = pred(s^*_v) \text{,}
\end{equation}
\begin{align*}
    \text{where } &s^*_v = \underset{s_v\in v}{\arg\max} \quad \{ \underset{c\in \mathcal{C}}{\max} \quad \%user\_votes_{s_v}(c) \} \text{,}\\
    &pred(s^*_v) = \underset{c\in \mathcal{C} 
   }{\arg\max} \quad \%user\_votes_{s^*_v}(c) \text{.}
\end{align*}

\section{Results}
\label{sec:results}

We include in our study a representative dataset from complex action recognition, Breakfast \cite{kuehne2014language}, one instructional video dataset, CrossTask \cite{zhukov2019cross}, and the Long-Form Video Understanding (LVU) dataset \cite{wu2021towards}.
We implement the user study on Amazon Mechanical Turk \cite{mturk} and collect responses from 167 users. We collect, on average, 12.09$\pm$1.62 votes for each video and video segment, which is proved to be a proper 
amount \cite{carvalho2016many}.
Table \ref{tab:accuracy_res} provides an overview of the results from the \textit{Full Videos Survey} and the \textit{Video Segments Survey}, discussed in the following sections. 

\begin{table}[!ht]
\centering
\begin{tabular}{l|cc}
\hline
\multirow{2}{*}{\textbf{Dataset}} & \multicolumn{2}{c}{\textbf{Classification accuracy (\%)}} \\
 & Full Videos & Video Segments \\
\midrule
\textbf{Breakfast} & \textbf{93.33} & 90.0 \\
\textbf{CrossTask} & \textbf{100.0} & 97.2 \\
\textbf{LVU – Relationship}& \textbf{88.89} & \textbf{88.89} \\
\textbf{LVU – Scene} & \textbf{100.0} & \textbf{100.0} \\
\textbf{LVU – Speaking} & \textbf{80.0} & 60.0 \\
\hline
\end{tabular}
\caption{Average video recognition accuracy obtained from the \textit{Full Videos Survey} and \textit{Video Segments Survey} on the Breakfast \cite{kuehne2014language}, CrossTask \cite{zhukov2019cross} and LVU \cite{wu2021towards} datasets. 
The results suggest that long-term information is helpful but not necessary in the majority of the evaluated datasets.}
\label{tab:accuracy_res}
\end{table}

\begin{table}[!ht]
\centering
\begin{tabular}{l|lll}
\hline 
\multirow{2}{*}{\textbf{Dataset}} & \multicolumn{3}{c}{\begin{tabular}[c]{@{}c@{}}\textbf{User agreement}\end{tabular}} \\
 & \begin{tabular}[c]{@{}c@{}}Full\\Videos\end{tabular} & \begin{tabular}[c]{@{}c@{}}Video\\Segments\end{tabular} & \begin{tabular}[c]{@{}c@{}}Selected\\Segments\end{tabular}\\
 \midrule 
\textbf{Breakfast} & \textbf{0.717} & 0.386 & 0.593 \\
\textbf{CrossTask} & 0.671 & 0.462 & 0.\textbf{767} \\
\textbf{LVU – relationship} & 0.499 & 0.340 & 0.\textbf{523} \\
\textbf{LVU – scene} & \textbf{0.755} & 0.481 & 0.686 \\
\textbf{LVU – speaking} & 0.159 & 0.191 & \textbf{0.265}\\
\hline
\end{tabular}
\caption{Overview of the user agreement in our user studies, measured terms of Krippendorff's $\alpha$ \cite{krippendorff2011computing}. 
We find that the users tend to agree in the \textit{Full Videos Surveys} and when selecting the segments with highest amount of votes for a class. Recognizing the actions in the \textit{Video Segments Survey} is generally harder then when looking at the full video, resulting in more variability in the users predictions and, consequently, in lower agreement.}
\label{tab:agreement}
\vspace*{-2mm}
\end{table}


\begin{figure*}[!ht]
\centering
\includegraphics[width=1\textwidth]{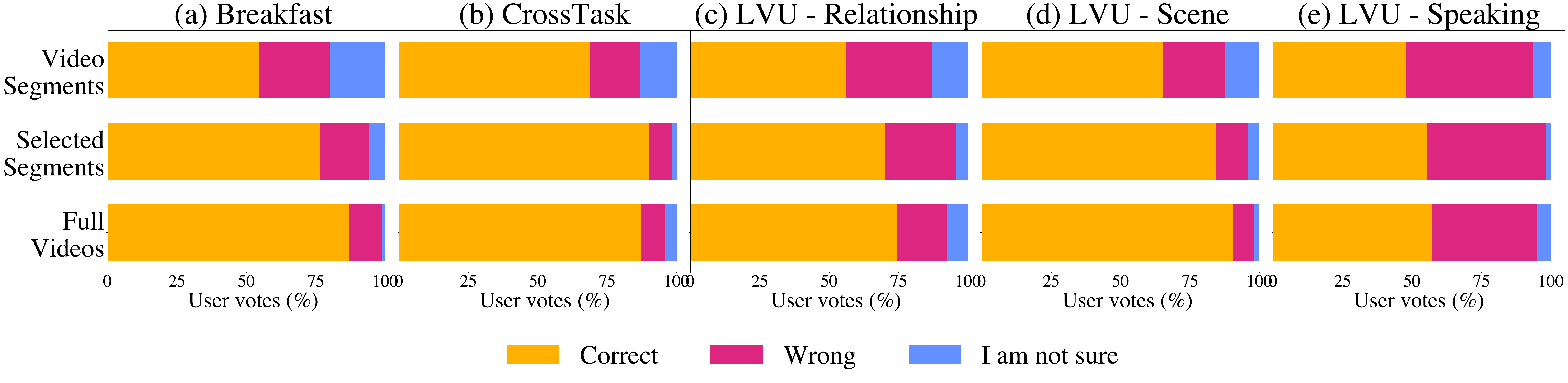}
\caption{
Overview of the user votes (correct, wrong and \textit{I am not sure}) collected in our study. We compare the results from the \textit{Full Videos}, all the \textit{Video Segments}, and the Selected Segments with highest percentage of votes for one class. 
%
The amount of correct votes in the Selected Segments is significantly higher than for all the \textit{Video Segments}, and comparable, or even higher, to the amount of correct votes obtained watching the full videos.
N.b., the user votes reported in this figure do not have to match the accuracies in Table \ref{tab:accuracy_res}. While the accuracy shows the percentage of videos correctly classified, the user votes are aggregated without considering the votes distributions within the specific videos.}
\label{fig:votes_res}
\end{figure*}

\subsection{Breakfast}

Breakfast \cite{kuehne2014language} is a collection of third-person videos of actors cooking a breakfast recipe, like scrambled eggs, coffee, cereals and milk. Each video has a global label, which corresponds to the recipe being made, for a total of 10 classes. The classification task consists in correctly recognizing the recipe.

For our study, we select a representative subset of 30 videos, corresponding to 3 randomly selected videos per class. The full videos have average duration of 2.44 $\pm$ 2.18 minutes. For the \textit{Video Segments Survey}, we segment the video according to the short-term action timesteps (\textit{coarse segmentation}) provided in the dataset. We remove 
segments that are shorter than 5 seconds, as we 
deem those segments highly uninformative, and we obtain 154 segments in total, of average duration 29 $\pm$ 39 seconds, where $\sim$56\% of the segments last less than 15 seconds. The large standard deviation is due to some 
repetitive short-term actions that can last above a minute, e.g. \textit{stir dough} or \textit{fry egg}.

The results in Table \ref{tab:accuracy_res} show that the recognition accuracy from the \textit{Full Videos Survey} (93.33\%) and the \textit{Video Segments Survey} (90.0\%) are close. This suggests that, although having access to the full long-term information in the video helps, looking at single short segments is sufficient to infer the right recipe class for the majority of the videos. From this result we conclude that the Breakfast dataset is not a proper long-term action dataset, according to our definition.

We analyze the amount of correct user votes, wrong votes and \textit{I am not sure} votes obtained in the user study and illustrated in Figure \ref{fig:votes_res} (a). We obtained 86.78\% of correct votes in the \textit{Full Videos Survey} and 54.47\% in the \textit{Videos Segments Survey}. However, if we consider only the segments with the highest percentage of votes for one class, 
the amount of correct votes reaches 
76.36\%. A similar trend occurs in the user agreement in Table \ref{tab:agreement}.
By further inspecting the results from the \textit{Video Segments Survey}, 
we notice that users are generally more uncertain classifying the video segments early in the video, with a higher portion of \textit{I am not sure} votes compare to the later segments. In particular, 63.57\% of \textit{I am not sure} votes are obtained in from the first two video segments in chronological order. We argue that breakfast dishes are usually 
better recognizable towards the end of the video, when the recipe is complete.

\subsection{CrossTask}

CrossTask \cite{zhukov2019cross} is an instructional video dataset of $\sim$4.7k videos, covering themes like auto repair, cooking and DIY.
The instructional videos show how to perform a \textit{tasks} (e.g., \textit{Make a Latte}) through a list of \textit{steps} (e.g., \textit{add coffee}, \textit{press coffee}, \textit{pour water}, \textit{pour espresso}, \textit{steam milk}, \textit{pour milk}). 
It contains 18 primary tasks with steps annotations and 65 related tasks with unlabeled steps. The dataset is meant to be used to learn steps in a weakly supervised learning setup. Here, we evaluate whether predicting the \textit{task} illustrated in an instructional video 
also fits our definition of long-term action recognition.
We collect results from 
36 video clips (2 random videos per primary task) 
of average duration 4.50 $\pm$ 2.14 minutes. Similarly to Breakfast, we extract 260 segments from the videos according to the 
timesteps provided with the dataset. In CrossTask, the segments are significantly shorter than Breakfast, with average duration of 10 $\pm$ 11 seconds and $\sim$81\% of the segments being shorter than 
15 seconds. 

In Table \ref{tab:accuracy_res}, we compare the task recognition accuracy from the \textit{Full Videos Survey}, 100\%, and the \textit{Video Segments Survey}, 97.2\%. In both cases, users can recognize the task with high accuracy. 
Only one video (YouTube id \href{https://www.youtube.com/watch?v=kReUYklvjnc}{\textit{kReUYklvjnc}}) is misclassified in the \textit{Video Segments Survey}, despite 5/8 of its video segments being correctly classified.
%
%
Considering the user agreement (Table \ref{tab:agreement}) and correct votes by the users (Figure \ref{fig:votes_res}, b), we find that both quantities are marginally higher in the Selected Segments over the Full Videos. This result shows that users tend to make the same mistakes (as for video \textit{kReUYklvjnc}) 
while confirming that most of the tasks are generally recognizable both from short video segments and full videos.
%
It is worth noting that the results reported in Table \ref{tab:accuracy_res} and Figure \ref{fig:votes_res} are not necessarily the same. The accuracy corresponds to the percentage of videos correctly classified, while the user votes are aggregated without considering the votes distributions within the specific videos.
Because of the high task recognition accuracy obtained from the \textit{Video Segments Survey}, we conclude that the videos in CrossTask do not contain long-term actions. We recommend to use this dataset for the other video understanding tasks that is supports, like captioning and action localization.


\subsection{LVU} 

The Long-Form Video Dataset (LVU) \cite{wu2021towards} has been recently proposed to study complex relationships in video clips extracted from movies. It provides three tasks, related to content understanding, user engagement prediction and movie metadata prediction and contains over 11k videos.
Similarly to previous work \cite{sun2022long}, we select the task of \textit{Content Understanding}, which involves classifying the \textit{relationship} among the characters, where the \textit{scene} is taking place and the characters \textit{speaking} style, from video clips of $\sim$2.5 minutes. 
The respective annotations consist in a global label per video. 
We assess whether predicting \textit{Relationship}, \textit{Scene} and \textit{Speaking} is a form of long-term action recognition, according to our definition.
We select videos from the test set and manually extract segments for each of the three classification tasks. We obtain 9 videos (3 per class) for \textit{Relationship}, 12 videos (2 per class) for \textit{Scene} and 10 videos (2 per class) for \textit{Speaking}, and a total of 140 segments of $\sim$30 seconds.

\begin{figure*}
     \centering
     \includegraphics[width=1\textwidth]{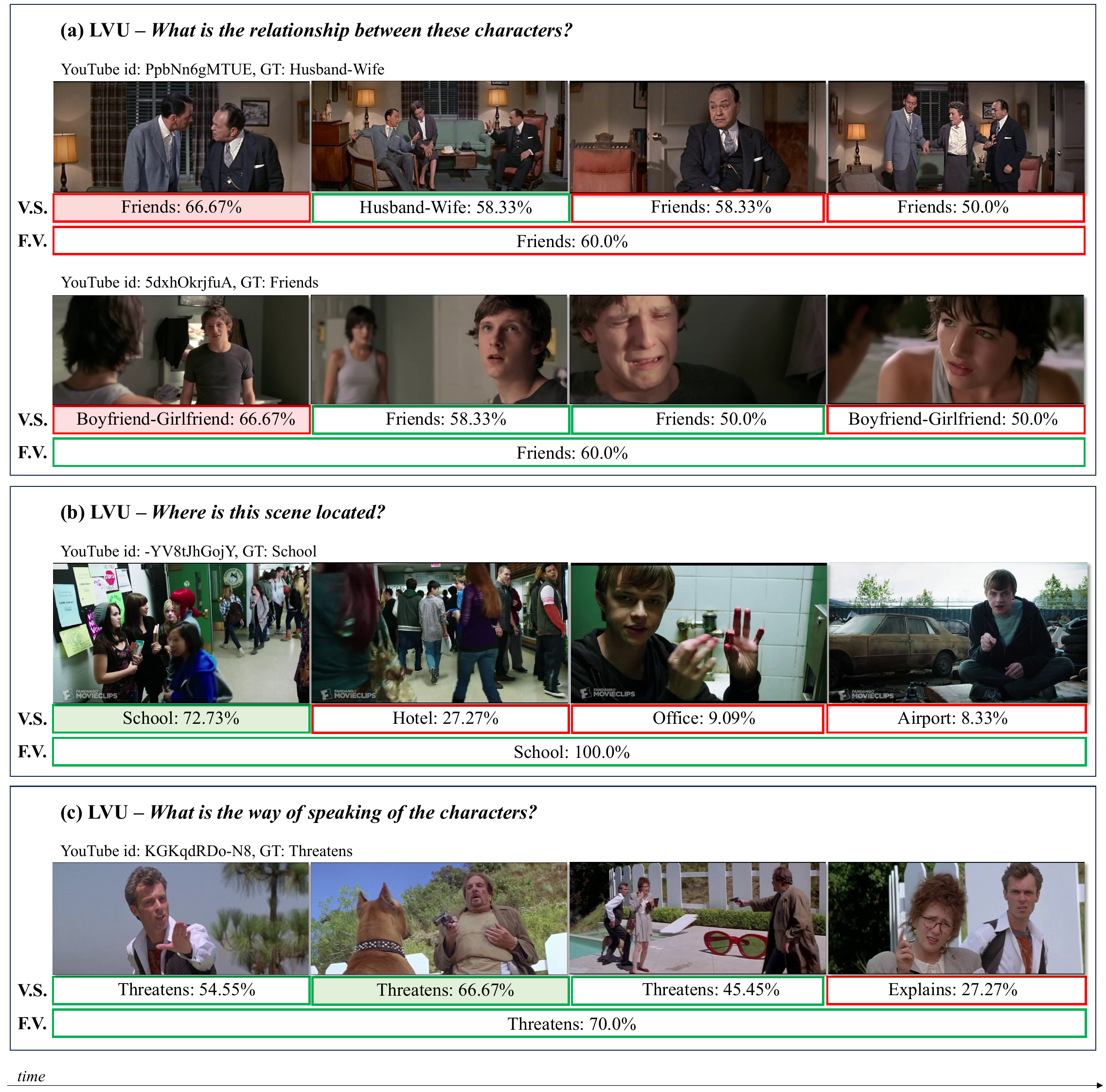}
    \caption{
    Examples of correct (green) and wrong (red) classification results collected from the \textit{Video Segments} (V.S.) and \textit{Full Videos} (F.V.) surveys on the Long-form Video Understanding (LVU) - Relationship (a), Scene (b) and Speaking (c) dataset \cite{wu2021towards}. Users correctly classify a large portion of video segments. Other segments result misclassified due to annotation noise.}
    \label{fig:LVU_Results}
\vspace*{-3mm}
\end{figure*}

Table \ref{tab:accuracy_res} shows the classification accuracies obtained from the \textit{Full Videos Survey} and \textit{Video Segments Survey}. Comparing the results, we find no difference for \textit{Relationship} and \textit{Scene}. In particular, \textit{Scene} classification is performed with 100\% accuracy, indicating that this prediction task is easy for humans. 
We identify a problem associated with LVU - \textit{Relationship}. The labels {husband-wife}, {friends}, {boyfriend-girlfriend} are associated with specific characters in the movie, but other characters might appear within the same video clip. For example, in Figure \ref{fig:LVU_Results} (a), the ground-truth label for the movie in the first row is \textit{Husband-Wife}. However, a third male character appears in the scene in addition to the \textit{husband and wife}. Therefore, the labels only correctly apply to a specific subset of the characters in the scene, or to a precise time window when only the target characters appear. 
As a result, the full videos are classified with a high percentage of wrong votes, while some of the video segments that do not include the characters corresponding to the label are completely misclassified. This justifies the large portion of wrong votes in Figure \ref{fig:votes_res} (c) and relatively low agreement in Table \ref{tab:agreement}.

We find 
a similar annotation problem in LVU - \textit{Speaking}. Also in this case, the global label only applies to a subset of the characters in the scene. In the example in Figure \ref{fig:LVU_Results} (c), the label \textit{Threatens} only applies to the man with the gun. This explains the difference in performance when comparing the accuracies from the \textit{Full Videos Survey} and \textit{Video Segments Survey} in Table \ref{tab:accuracy_res}, the large amount of wrong votes in Figure \ref{fig:votes_res} (e) and low agreement in Table \ref{tab:agreement}.
Because of the problem with the annotations and the equal recognition performance of 88.89\% obtained from the \textit{Full Videos Survey} and \textit{Video Segments Survey} (reported in Table \ref{tab:accuracy_res}), we conclude that LVU - \textit{Relationship} is not a long-term video understanding task. Similar conclusions apply for LVU - \textit{Scene}, with perfect classification scores resulting from both surveys. Finally, the labels in LVU - \textit{Speaking} are not truly long-term, as they apply to a subset of characters speaking only during some relatively short time-windows.

\section{Conclusion}
We propose a method to assess whether an action is \textit{long-term}. We apply our method to three current long-term video understanding datasets, Breakfast, CrossTask and LVU. Our results show that long-term information 
might help but is \textit{not necessary} in the majority of videos from the analyzed datasets. 
In fact, the long-term actions in these videos can be correctly classified by humans by looking solely at a single short video segment.
This result suggests that deep learning models trained and tested on these datasets might pick short-term shortcuts and still show correct recognition performance, without actually learning any long-term information. 
Following our findings, we urge researchers who are investigating automatic long-term action recognition to use 
datasets that need long-term information to be solved. 


\small
\smallskip\noindent\textbf{Acknowledgements.} 
This work is part of the research program Efficient Deep Learning (EDL), which is (partly) financed by the Dutch Research Council (NWO).

{\small
\bibliographystyle{ieee_fullname}
\bibliography{egbib}

\begin{thebibliography}{10}\itemsep=-1pt

\bibitem{mturk}
Amazon mechanical turk.
\newblock \url{https://www.mturk.com/}.
\newblock Accessed: 2023-07-05.

\bibitem{alayrac2017learning}
Jean-Baptiste Alayrac, Piotr Bojanowski, Nishant Agrawal, Josef Sivic, Ivan
  Laptev, and Simon Lacoste-Julien.
\newblock Learning from narrated instruction videos.
\newblock {\em IEEE transactions on pattern analysis and machine intelligence},
  40(9):2194--2208, 2017.

\bibitem{arnab2021vivit}
Anurag Arnab, Mostafa Dehghani, Georg Heigold, Chen Sun, Mario Lu{\v{c}}i{\'c},
  and Cordelia Schmid.
\newblock Vivit: A video vision transformer.
\newblock In {\em Proceedings of the IEEE/CVF International Conference on
  Computer Vision}, pages 6836--6846, 2021.

\bibitem{ballan2021long}
Luca Ballan, Ombretta Strafforello, and Klamer Schutte.
\newblock Long-term behaviour recognition in videos with actor-focused region
  attention.
\newblock In {\em VISIGRAPP (5: VISAPP)}, pages 362--369, 2021.

\bibitem{bertasius2021space}
Gedas Bertasius, Heng Wang, and Lorenzo Torresani.
\newblock Is space-time attention all you need for video understanding?
\newblock In {\em ICML}, volume~2, page~4, 2021.

\bibitem{byvshev20223d}
Petr Byvshev, Pascal Mettes, and Yu Xiao.
\newblock Are 3d convolutional networks inherently biased towards appearance?
\newblock {\em Computer Vision and Image Understanding}, 220:103437, 2022.

\bibitem{caba2015activitynet}
Fabian Caba~Heilbron, Victor Escorcia, Bernard Ghanem, and Juan Carlos~Niebles.
\newblock Activitynet: A large-scale video benchmark for human activity
  understanding.
\newblock In {\em Proceedings of the ieee Conference on Computer Vision and
  Pattern Recognition}, pages 961--970, 2015.

\bibitem{carreira2017quo}
Joao Carreira and Andrew Zisserman.
\newblock Quo vadis, action recognition? a new model and the kinetics dataset.
\newblock In {\em proceedings of the IEEE Conference on Computer Vision and
  Pattern Recognition}, pages 6299--6308, 2017.

\bibitem{carvalho2016many}
Arthur Carvalho, Stanko Dimitrov, and Kate Larson.
\newblock How many crowdsourced workers should a requester hire?
\newblock {\em Annals of Mathematics and Artificial Intelligence}, 78:45--72,
  2016.

\bibitem{choi2019can}
Jinwoo Choi, Chen Gao, Joseph~CE Messou, and Jia-Bin Huang.
\newblock Why can't i dance in the mall? learning to mitigate scene bias in
  action recognition.
\newblock {\em Advances in Neural Information Processing Systems}, 32, 2019.

\bibitem{damen2018scaling}
Dima Damen, Hazel Doughty, Giovanni~Maria Farinella, Sanja Fidler, Antonino
  Furnari, Evangelos Kazakos, Davide Moltisanti, Jonathan Munro, Toby Perrett,
  Will Price, et~al.
\newblock Scaling egocentric vision: The epic-kitchens dataset.
\newblock In {\em Proceedings of the European Conference on Computer Vision
  (ECCV)}, pages 720--736, 2018.

\bibitem{feichtenhofer2019slowfast}
Christoph Feichtenhofer, Haoqi Fan, Jitendra Malik, and Kaiming He.
\newblock Slowfast networks for video recognition.
\newblock In {\em Proceedings of the IEEE/CVF International Conference on
  Computer Vision}, pages 6202--6211, 2019.

\bibitem{geirhos2020shortcut}
Robert Geirhos, J{\"o}rn-Henrik Jacobsen, Claudio Michaelis, Richard Zemel,
  Wieland Brendel, Matthias Bethge, and Felix~A Wichmann.
\newblock Shortcut learning in deep neural networks.
\newblock {\em Nature Machine Intelligence}, 2(11):665--673, 2020.

\bibitem{girdhar2019cater}
Rohit Girdhar and Deva Ramanan.
\newblock Cater: A diagnostic dataset for compositional actions and temporal
  reasoning.
\newblock {\em arXiv preprint arXiv:1910.04744}, 2019.

\bibitem{goyal2017something}
Raghav Goyal, Samira Ebrahimi~Kahou, Vincent Michalski, Joanna Materzynska,
  Susanne Westphal, Heuna Kim, Valentin Haenel, Ingo Fruend, Peter Yianilos,
  Moritz Mueller-Freitag, et~al.
\newblock The" something something" video database for learning and evaluating
  visual common sense.
\newblock In {\em Proceedings of the IEEE International Conference on Computer
  Vision}, pages 5842--5850, 2017.

\bibitem{guo2022uncertainty}
Hongji Guo, Hanjing Wang, and Qiang Ji.
\newblock Uncertainty-guided probabilistic transformer for complex action
  recognition.
\newblock In {\em Proceedings of the IEEE/CVF Conference on Computer Vision and
  Pattern Recognition}, pages 20052--20061, 2022.

\bibitem{hussein2019timeception}
Noureldien Hussein, Efstratios Gavves, and Arnold~WM Smeulders.
\newblock Timeception for complex action recognition.
\newblock In {\em Proceedings of the IEEE/CVF Conference on Computer Vision and
  Pattern Recognition}, pages 254--263, 2019.

\bibitem{hussein2019videograph}
Noureldien Hussein, Efstratios Gavves, and Arnold~WM Smeulders.
\newblock Videograph: Recognizing minutes-long human activities in videos.
\newblock {\em arXiv preprint arXiv:1905.05143}, 2019.

\bibitem{hussein2020timegate}
Noureldien Hussein, Mihir Jain, and Babak~Ehteshami Bejnordi.
\newblock Timegate: Conditional gating of segments in long-range activities.
\newblock {\em arXiv preprint arXiv:2004.01808}, 2020.

\bibitem{islam2022long}
Md~Mohaiminul Islam and Gedas Bertasius.
\newblock Long movie clip classification with state-space video models.
\newblock In {\em Proceedings of the European Conference on Computer Vision
  (ECCV)}, pages 87--104. Springer, 2022.

\bibitem{ji2020action}
Jingwei Ji, Ranjay Krishna, Li Fei-Fei, and Juan~Carlos Niebles.
\newblock Action genome: Actions as compositions of spatio-temporal scene
  graphs.
\newblock In {\em Proceedings of the IEEE/CVF Conference on Computer Vision and
  Pattern Recognition}, pages 10236--10247, 2020.

\bibitem{ji20123d}
Shuiwang Ji, Wei Xu, Ming Yang, and Kai Yu.
\newblock 3d convolutional neural networks for human action recognition.
\newblock {\em IEEE transactions on pattern analysis and machine intelligence},
  35(1):221--231, 2012.

\bibitem{krippendorff2011computing}
Klaus Krippendorff.
\newblock Computing krippendorff's alpha-reliability.
\newblock {\em Departmental Papers (ASC), University of Pennsylvania}, 2011.

\bibitem{kuehne2014language}
Hilde Kuehne, Ali Arslan, and Thomas Serre.
\newblock The language of actions: Recovering the syntax and semantics of
  goal-directed human activities.
\newblock In {\em Proceedings of the IEEE Conference on Computer Vision and
  Pattern Recognition}, pages 780--787, 2014.

\bibitem{kuehne2011hmdb}
Hildegard Kuehne, Hueihan Jhuang, Est{\'\i}baliz Garrote, Tomaso Poggio, and
  Thomas Serre.
\newblock Hmdb: a large video database for human motion recognition.
\newblock In {\em 2011 International Conference on Computer Vision}, pages
  2556--2563. IEEE, 2011.

\bibitem{li2022bridge}
Muheng Li, Lei Chen, Yueqi Duan, Zhilan Hu, Jianjiang Feng, Jie Zhou, and Jiwen
  Lu.
\newblock Bridge-prompt: Towards ordinal action understanding in instructional
  videos.
\newblock In {\em Proceedings of the IEEE/CVF Conference on Computer Vision and
  Pattern Recognition}, pages 19880--19889, 2022.

\bibitem{liu2022video}
Ze Liu, Jia Ning, Yue Cao, Yixuan Wei, Zheng Zhang, Stephen Lin, and Han Hu.
\newblock Video swin transformer.
\newblock In {\em Proceedings of the IEEE/CVF Conference on Computer Vision and
  Pattern Recognition}, pages 3202--3211, 2022.

\bibitem{miech2020end}
Antoine Miech, Jean-Baptiste Alayrac, Lucas Smaira, Ivan Laptev, Josef Sivic,
  and Andrew Zisserman.
\newblock End-to-end learning of visual representations from uncurated
  instructional videos.
\newblock In {\em Proceedings of the IEEE/CVF Conference on Computer Vision and
  Pattern Recognition}, pages 9879--9889, 2020.

\bibitem{rohrbach2012script}
Marcus Rohrbach, Michaela Regneri, Mykhaylo Andriluka, Sikandar Amin, Manfred
  Pinkal, and Bernt Schiele.
\newblock Script data for attribute-based recognition of composite activities.
\newblock In {\em Proceedings of the European Conference on Computer Vision
  (ECCV)}, pages 144--157. Springer, 2012.

\bibitem{sharghi2020automatic}
Aidean Sharghi, Helene Haugerud, Daniel Oh, and Omid Mohareri.
\newblock Automatic operating room surgical activity recognition for
  robot-assisted surgery.
\newblock In {\em Medical Image Computing and Computer Assisted
  Intervention--MICCAI 2020: 23rd International Conference, Lima, Peru, October
  4--8, 2020, Proceedings, Part III 23}, pages 385--395. Springer, 2020.

\bibitem{sigurdsson2016hollywood}
Gunnar~A Sigurdsson, G{\"u}l Varol, Xiaolong Wang, Ali Farhadi, Ivan Laptev,
  and Abhinav Gupta.
\newblock Hollywood in homes: Crowdsourcing data collection for activity
  understanding.
\newblock In {\em Proceedings of the European Conference on Computer Vision
  (ECCV)}, pages 510--526. Springer, 2016.

\bibitem{soomro2012ucf101}
Khurram Soomro, Amir~Roshan Zamir, and Mubarak Shah.
\newblock Ucf101: A dataset of 101 human actions classes from videos in the
  wild.
\newblock {\em arXiv preprint arXiv:1212.0402}, 2012.

\bibitem{stein2013combining}
Sebastian Stein and Stephen~J McKenna.
\newblock Combining embedded accelerometers with computer vision for
  recognizing food preparation activities.
\newblock In {\em Proceedings of the 2013 ACM international joint conference on
  Pervasive and ubiquitous computing}, pages 729--738, 2013.

\bibitem{videobagnet}
Ombretta Strafforello, Xin Liu, Klamer Schutte, and Jan van Gemert.
\newblock Video bagnet: short temporal receptive fields increase robustness in
  long-term action recognition.
\newblock In {\em Proceedings of the IEEE/CVF International Conference on
  Computer Vision Workshops}, 2023.

\bibitem{sun2022long}
Yuchong Sun, Hongwei Xue, Ruihua Song, Bei Liu, Huan Yang, and Jianlong Fu.
\newblock Long-form video-language pre-training with multimodal temporal
  contrastive learning.
\newblock {\em arXiv preprint arXiv:2210.06031}, 2022.

\bibitem{tang2019coin}
Yansong Tang, Dajun Ding, Yongming Rao, Yu Zheng, Danyang Zhang, Lili Zhao,
  Jiwen Lu, and Jie Zhou.
\newblock Coin: A large-scale dataset for comprehensive instructional video
  analysis.
\newblock In {\em Proceedings of the IEEE/CVF Conference on Computer Vision and
  Pattern Recognition}, pages 1207--1216, 2019.

\bibitem{toyer2017human}
Sam Toyer, Anoop Cherian, Tengda Han, and Stephen Gould.
\newblock Human pose forecasting via deep markov models.
\newblock In {\em 2017 International Conference on Digital Image Computing:
  Techniques and Applications (DICTA)}, pages 1--8. IEEE, 2017.

\bibitem{tran2015learning}
Du Tran, Lubomir Bourdev, Rob Fergus, Lorenzo Torresani, and Manohar Paluri.
\newblock Learning spatiotemporal features with 3d convolutional networks.
\newblock In {\em Proceedings of the IEEE International Conference on Computer
  Vision}, pages 4489--4497, 2015.

\bibitem{7961782}
Geoffrey Vaquette, Astrid Orcesi, Laurent Lucat, and Catherine Achard.
\newblock The daily home life activity dataset: A high semantic activity
  dataset for online recognition.
\newblock In {\em 2017 12th IEEE International Conference on Automatic Face \&
  Gesture Recognition (FG 2017)}, pages 497--504, 2017.

\bibitem{wang2016temporal}
Limin Wang, Yuanjun Xiong, Zhe Wang, Yu Qiao, Dahua Lin, Xiaoou Tang, and Luc
  Van~Gool.
\newblock Temporal segment networks: Towards good practices for deep action
  recognition.
\newblock In {\em Proceedings of the European Conference on Computer Vision
  (ECCV)}, pages 20--36. Springer, 2016.

\bibitem{wu2021towards}
Chao-Yuan Wu and Philipp Krahenbuhl.
\newblock Towards long-form video understanding.
\newblock In {\em Proceedings of the IEEE/CVF Conference on Computer Vision and
  Pattern Recognition}, pages 1884--1894, 2021.

\bibitem{wu2022memvit}
Chao-Yuan Wu, Yanghao Li, Karttikeya Mangalam, Haoqi Fan, Bo Xiong, Jitendra
  Malik, and Christoph Feichtenhofer.
\newblock Memvit: Memory-augmented multiscale vision transformer for efficient
  long-term video recognition.
\newblock In {\em Proceedings of the IEEE/CVF Conference on Computer Vision and
  Pattern Recognition}, pages 13587--13597, 2022.

\bibitem{xiao2021noise}
Kai~Yuanqing Xiao, Logan Engstrom, Andrew Ilyas, and Aleksander Madry.
\newblock Noise or signal: The role of image backgrounds in object recognition.
\newblock In {\em International Conference on Learning Representations}, 2021.

\bibitem{yang2023relational}
Xitong Yang, Fu-Jen Chu, Matt Feiszli, Raghav Goyal, Lorenzo Torresani, and Du
  Tran.
\newblock Relational space-time query in long-form videos.
\newblock In {\em Proceedings of the IEEE/CVF Conference on Computer Vision and
  Pattern Recognition}, pages 6398--6408, 2023.

\bibitem{yeung2018every}
Serena Yeung, Olga Russakovsky, Ning Jin, Mykhaylo Andriluka, Greg Mori, and Li
  Fei-Fei.
\newblock Every moment counts: Dense detailed labeling of actions in complex
  videos.
\newblock {\em International Journal of Computer Vision}, 126:375--389, 2018.

\bibitem{yu2020rhyrnn}
Tianshu Yu, Yikang Li, and Baoxin Li.
\newblock Rhyrnn: Rhythmic rnn for recognizing events in long and complex
  videos.
\newblock In {\em Proceedings of the European Conference on Computer Vision
  (ECCV)}, pages 127--144. Springer, 2020.

\bibitem{zhou2021graph}
Jiaming Zhou, Kun-Yu Lin, Haoxin Li, and Wei-Shi Zheng.
\newblock Graph-based high-order relation modeling for long-term action
  recognition.
\newblock In {\em Proceedings of the IEEE/CVF Conference on Computer Vision and
  Pattern Recognition}, pages 8984--8993, 2021.

\bibitem{zhou2018towards}
Luowei Zhou, Chenliang Xu, and Jason Corso.
\newblock Towards automatic learning of procedures from web instructional
  videos.
\newblock In {\em Proceedings of the AAAI Conference on Artificial
  Intelligence}, volume~32, 2018.

\bibitem{zhukov2019cross}
Dimitri Zhukov, Jean-Baptiste Alayrac, Ramazan~Gokberk Cinbis, David Fouhey,
  Ivan Laptev, and Josef Sivic.
\newblock Cross-task weakly supervised learning from instructional videos.
\newblock In {\em Proceedings of the IEEE/CVF Conference on Computer Vision and
  Pattern Recognition}, pages 3537--3545, 2019.

\end{thebibliography}
}

\end{document}